\documentclass[runningheads]{llncs}
\usepackage[T1]{fontenc}
\usepackage{graphicx}
\usepackage{booktabs}
\usepackage[misc]{ifsym}
\usepackage{hyperref}
\usepackage{url}
\usepackage{pifont}
\usepackage{adjustbox}
\usepackage{tabularx}
\usepackage{placeins}
\usepackage{enumitem}
\usepackage{booktabs}
\usepackage{caption}
\usepackage{lipsum}
\usepackage{wrapfig}
\usepackage{listings}
\usepackage{upquote}
\usepackage{diagbox}
\usepackage{multirow}
\usepackage{tikz}
\usepackage{array}
\usepackage{xcolor}
\usepackage{xr}
\usepackage[utf8]{inputenc}

\newcommand{\bigslashcell}[1][3cm]{%
  \begin{tikzpicture}
    \draw (0,0) -- (#1,2.3cm);        
  \end{tikzpicture}%
}

\usepackage{mwe}

\begin{document}

\title{Data Analysis in the Wild: Benchmarking Large Language Models Against Real-World Data Complexities}

\titlerunning{Data Analysis in the Wild}

\author{So Hasegawa\inst{1} \and
Shailaja Keyur Sampat\inst{1} \and
Lei Liu\inst{1} \and
Wei-Peng Chen\inst{1}
}

\authorrunning{So Hasegawa et al.}

\institute{Fujitsu Research of America, Santa Clara CA 95054, USA \email{shasegawa@fujitsu.com}}

\maketitle              

\newcommand{\cmark}{\ding{51}} 
\newcommand{\xmark}{\ding{55}} 

\newcommand*\colourcheck[1]{%
  \expandafter\newcommand\csname #1check\endcsname{\textcolor{#1}{\ding{51}}}%
}
\colourcheck{blue}
\colourcheck{green}
\colourcheck{red}

\newcommand{\fix}{\marginpar{FIX}}
\newcommand{\new}{\marginpar{NEW}}


\begin{abstract}
Current benchmarks for evaluating Large Language Models (LLMs) in data analysis often fail to reflect real-world settings. They typically focus on fact retrieval from small tables and overlook the challenges of large multi-tabular datasets, external knowledge integration, and exploratory insight discovery.
We introduce \textit{DataGovBench}, a benchmark derived from governmental open data designed to evaluate LLMs in practical scenarios. The benchmark includes two tasks: \textit{Table QA} that requires solving complex decomposable questions and producing textual answers or visualizations, and \textit{Table Insight} that evaluates the ability of models to generate expert-level findings through exploratory data analysis.
Comprehensive experiments with state-of-the-art LLMs, both with and without agentic frameworks, reveal significant performance gaps across both tasks. These results suggest that current LLM-based systems remain far from satisfying the demands of real-world data analytics. DataGovBench provides a challenging benchmark for advancing research on LLMs capable of both answering analytical queries and discovering insights from data. Code and sample data are available at \url{https://github.com/SoHasegawa/datagovbench}.

\end{abstract}

\section{Introduction}

\begin{figure*}[t!]
\centering
\includegraphics[width=1.0\linewidth]{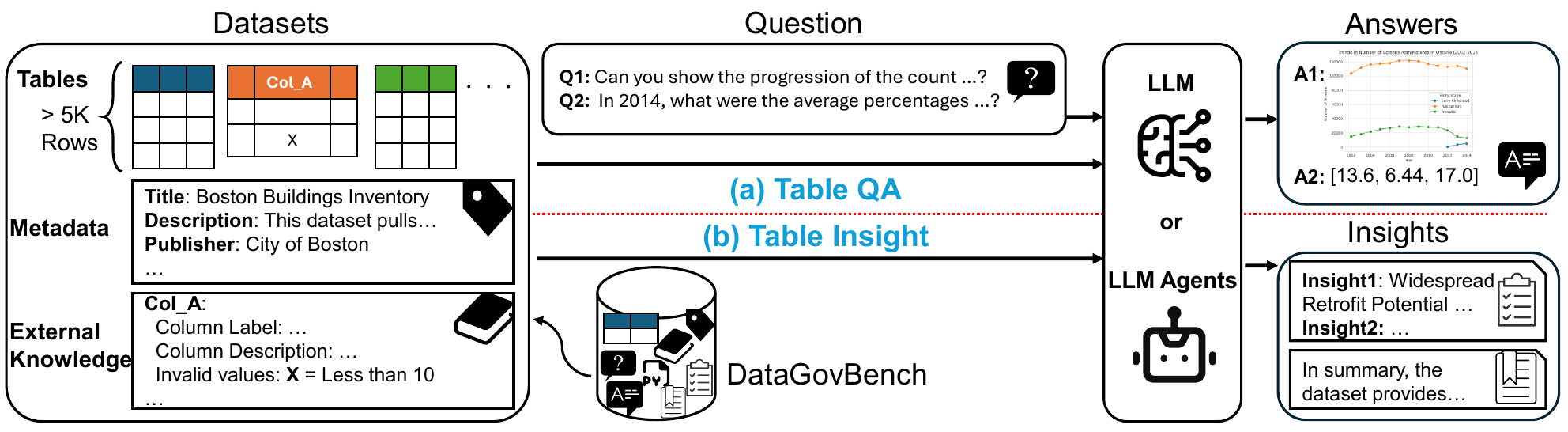}
\caption{DataGovBench evaluates LLMs and agents on two table reasoning tasks using large, multi-table datasets supplemented with metadata and external knowledge. (a) The \textit{Table QA} task requires models to answer simple or decomposable questions with textual or visual answers. (b) In contrast, the \textit{Table Insight} task challenges models to perform open-ended exploratory analysis, proactively generating a list of insights and a summary without a specific user query.}
\label{fig:user_diagram}
\end{figure*}

The ability to reason over structured data is a cornerstone of modern data science and a long-standing challenge in artificial intelligence. With the advent of Large Language Models (LLMs), we have witnessed a paradigm shift in how humans interact with complex information~\cite{openai2024gpt4technicalreport,team2023gemini,yang2025qwen3technicalreport,deepseekai2025deepseekr1incentivizingreasoningcapability}. These models have led to the development of sophisticated agents designed to democratize data analysis, promising a future where any user can pose natural language questions to a dataset and receive accurate answers~\cite{infida-bench,su2024tablegpt2largemultimodalmodel}. The ultimate vision is an autonomous system that not only retrieves information but also uncovers the knowledge hidden within raw data.

However, a significant gap persists between this vision and reality, as the benchmarks lack real-world complexity. While datasets like WikiTableQuestions~\cite{wtq} and Spider~\cite{yu-etal-2018-spider} propelled research in semantic parsing and text-to-SQL, their controlled environments use small-scale tables. They largely neglect practical challenges such as massive table scales, the need to merge multiple tables, and the essential role of metadata and external knowledge. 
Beyond these data limitations, existing benchmarks have focused on direct fact retrieval~\cite{wu2025tablebench,infida-bench,zhang-etal-2025-t2r}. Tasks like QA and text-to-SQL are about retrieving information to a query, while missing the capability of proactive insight discovery that data analysts exhibit. Consequently, this discovery-oriented skill remains largely unevaluated, as few benchmarks have formalized insight generation as a primary task~\cite{insight-bench,mt-raig}.

To bridge these notable gaps in both data realism and task scope, we introduce \textit{DataGovBench}, a comprehensive benchmark sourced from public repositories like Data.gov~\cite{datagov}. Our benchmark features two complementary tasks: Table Question Answering (\textit{Table QA}) and Table Insight Generation (\textit{Table Insight}), as illustrated in Figure~\ref{fig:user_diagram}. The Table QA task assesses factual reasoning over decomposable questions that explicitly include multiple sub-questions, requiring models to produce answers either in text or in visualizations. In contrast, the Table Insight task challenges models to perform expert-level insight derivation, requiring in-depth analysis and the discovery of trends.

For the Table QA task, we use LLMs, aided by a novel table serialization, to generate a diverse corpus of QA pairs that are then meticulously verified by human annotators. For the Table Insight task, we address the challenge of subjectivity by using the expert-authored reports accompanying datasets as a ground truth. This process yields a novel benchmark that surpasses existing alternatives by holistically combining the complex Table QA and Table Insight with the diverse data complexities, such as large-scale, multi-tabular datasets that require metadata and external knowledge, as detailed in Table~\ref{tab:comparison_table}.


Comprehensive evaluation unearths that latest top-performing LLMs achieve low accuracy for both tasks even with the agentic support. These results demonstrate that DataGovBench captures real-world challenges. Furthermore, we provide detailed qualitative analyses for both tasks, identifying common failure modes and uncovering two missing capabilities of current LLM agents: narrative-level reasoning over tabular data and accurate fact retrieval from complex tables.

In summary, our paper makes the following three contributions:
\begin{itemize}[leftmargin=*]
    \item We introduce \textit{DataGovBench}, featuring tasks for both multifaceted Question Answering and Insight Generation with ground-truths annotated by domain experts. These tasks comprehensively handle large, multi-tabular, and heterogeneous datasets representing complexity in real-world scenarios. 
    \item Extensive evaluation of state-of-the-art models reveals a crucial performance gap, as even top models achieve low accuracy despite agentic support. This underscores the benchmark's difficulty and its alignment with actual challenges.
    \item We provide a detailed qualitative analysis and ablation studies that identify key failure modes, offering a clear guide for the community to focus on primary areas for LLM-based data analytics.
\end{itemize}

\begin{table}[t!]
\centering
\footnotesize
\begin{adjustbox}{max width=0.95\textwidth}
\begin{tabular}{l|c|cc|cccc}
\toprule
\multicolumn{1}{c|}{\textbf{Benchmark}} & 
\multicolumn{1}{c|}{\textbf{Insight}} &  
\multicolumn{2}{c|}{\textbf{QA}} &
\multicolumn{4}{c}{\textbf{Dataset Characteristics}} \\
& & \textbf{Decomposable QA} & \textbf{Visualization} & 
\textbf{Large-table} & \textbf{Multiple Tables} & \textbf{Metadata} & \textbf{External Knowledge} \\
\midrule
\multicolumn{8}{c}{\textbf{Existing Benchmarks for Table Question and Answering}} \\
\midrule 
WTQ~\cite{wtq} & \xmark & \xmark & \xmark & \xmark & \xmark & \xmark & \xmark \\
OTT-QA~\cite{ottqa} & \xmark & \xmark & \xmark & \xmark & \greencheck & \greencheck & \greencheck \\
FeTaQA~\cite{nan-etal-2022-fetaqa} & \xmark & \xmark & \xmark & \xmark & \xmark & \greencheck & \xmark \\
DataBench~\cite{oses-grijalba-etal-2024-question} & \xmark & \xmark & \xmark & \greencheck & \xmark & \xmark & \xmark \\
TableBench~\cite{wu2025tablebench} & \xmark  & \xmark & \greencheck & \xmark & \xmark & \xmark & \xmark \\
MMQA~\cite{wu2025mmqa} & \xmark & \xmark & \xmark & \xmark & \greencheck & \xmark & \xmark \\ 
\midrule
\multicolumn{8}{c}{\textbf{Existing Benchmarks for Table Insight Generation}} \\
\midrule
InsightBench~\cite{insight-bench} & \greencheck & \xmark & \xmark & \xmark & \greencheck & \xmark & \xmark \\
MT-RAIG~\cite{mt-raig} & \greencheck & \xmark & \xmark & \xmark & \greencheck & \greencheck & \xmark \\
T2R-Bench~\cite{zhang-etal-2025-t2r} & \greencheck & \xmark & \xmark & \greencheck & \greencheck & \xmark & \xmark \\
\midrule
\textbf{DataGovBench} & \greencheck & \greencheck & \greencheck & \greencheck & \greencheck & \greencheck & \greencheck \\
\bottomrule
\end{tabular}
\end{adjustbox}
\caption{Comparison of existing Table QA and Table Insight benchmarks with respect to task coverage and dataset characteristics.}
\label{tab:comparison_table}
\end{table}

\section{Overview of DataGovBench}

DataGovBench is a new benchmark designed to evaluate tabular reasoning in realistic scenarios. The data is derived from governmental open data portals (e.g., Data.gov~\cite{datagov}, Data.gov.uk~\cite{datagovuk}), which host publicly available datasets from official institutions. Datasets on these portals reflect real-world complexity, consisting of a single or multiple tables containing a large number of records, and rich contextual information. This context is provided through metadata, such as textual descriptions of the dataset, and often supplemented with external knowledge like data dictionaries.
The benchmark is designed to evaluate performance on two core data science tasks: Table QA and Table Insight.

\textbf{Table QA}: This task requires models and agents to answer simple or decomposable questions with multiple sub-questions. The answers are provided in text or visualizations.

\textbf{Table Insight}: This task challenges models and agents to perform exploratory analysis, generating substantive insights directly from the tabular data without an explicit user query.

\subsection{Benchmark Construction}
\label{main_benchmark_construction}

\begin{figure*}[t!]
\centering
\includegraphics[width=0.95\linewidth]{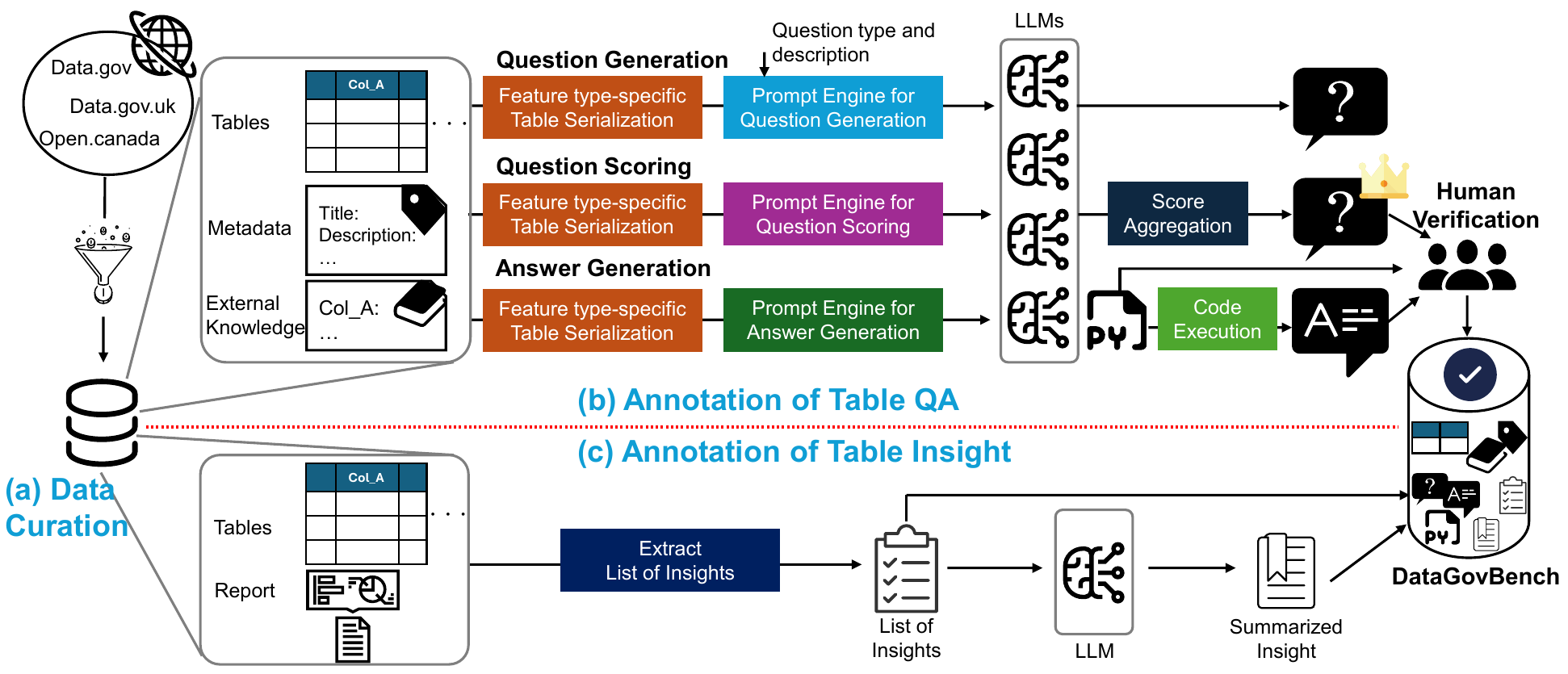}
\caption{Overview of the three-stage construction process of DataGovBench}
\label{fig:benchmark_construction}
\end{figure*}

The construction of DataGovBench involved a three-stage process as shown in Figure~\ref{fig:benchmark_construction}: 1) curating datasets from open data portals through a systematic filtering process, 2) annotating QA pairs via a human-in-the-loop, and 3) compiling ground-truth insights by leveraging professional reports. 

\subsubsection{Data Curation}
\label{data_curation}

Given the vast and decentralized nature of open data available online, a systematic collection and filtering process was imperative. Our process began by identifying 53 open data platforms with English as the primary language (a complete list is provided in the Table~\ref{tab:websites}).
We downloaded all available datasets from these platforms and then applied a series of filtering criteria including at least one of the tabular files covering over 5,000 records and metadata with description to understand the context of the dataset. A more detailed filtering procedure is outlined in the Appendix~\ref{sec:data_filtering}.

\subsubsection{Annotation of Question and Answer Pairs}
\label{annotation_qa}

To construct a high-quality, complex, and challenging set of QA pairs at scale while mitigating the need for resource-intensive manual annotation, we designed a four-stage generation pipeline that leverages LLMs with human-in-the-loop verification. Inspired by prior work~\cite{wu2025tablebench}, this approach ensures both diversity and correctness. The procedure is detailed below.
\begin{enumerate}[leftmargin=*]
    \item \textbf{Question Generation}: The initial stage focused on generating a diverse pool of candidate questions. To guide the output of the LLM, we first defined eight question types as presented in Appendix~\ref{question_types}. These types encompassed not only simple queries (e.g. Ranking, Aggregation) shared with existing benchmarks~\cite{wu2025tablebench,wu2025mmqa} but also complex decomposable questions. The prompt contains the table contents, the title and description of the dataset from the metadata, external knowledge when available, and the designated question type. 
    Generating diverse and meaningful questions with LLMs requires providing them with a representative view of the table contents and value distributions. While prior works mainly focus on smaller datasets such that the entire table or the first several rows could be embedded into the prompt~\cite{wu2025tablebench}, we cannot apply similar approaches to tables with millions of records in our benchmark as it exceeds the limits of LLM context window. To address this challenge, we propose a technique called \textit{Feature type-specific table serialization}, which creates a compact yet informative summary of a table by representing each column according to its data type. For instance, instead of listing all values in a categorical column, we provide only the set of unique categories. This serialization is a core component of our workflow, and a detailed description of the logic for various feature types is presented in Appendix~\ref{ft_serialization}.
    To mitigate model-specific biases in the generated questions, we employed an ensemble of four high-performance LLMs: GPT-4o, GPT-4o-mini~\cite{openai2024gpt4ocard}, Gemini 2.0 Flash~\cite{gemini2.0}, and Gemini 1.5 Pro~\cite{gemini15unlockingmutimodal}. 
    \item \textbf{Question Scoring}: Following generation, the candidate questions underwent an automated scoring and selection phase. Each question was evaluated against four qualitative criteria: relevance to the dataset, its potential to yield insightful or actionable information, sufficient analytical complexity with novelty, and clarity expressed in natural language. In an approach to quality filtering, we tasked each of the four aforementioned LLMs with acting as a judge, selecting its top-5 preferred questions from the generated pool based on the above criteria for each dataset. A score from 5 (highest) to 1 (lowest) was assigned to these selections. The scores from all four models were then aggregated for each question, resulting in a maximum possible score of 20. Based on this aggregated score, we selected the top-10 questions to proceed to the next stage.
    \item \textbf{Answer Generation}: For each of the top-10 questions per dataset, an LLM was prompted to generate Python code that produces the correct answer. After executing the code from all four models, we measured the answer consensus to filter out questions that yielded unanimous agreement across all four LLMs. Such instances were deemed to indicate a low level of analytical complexity (e.g. single column filtering or aggregation), making them unsuitable for a benchmark in a real-world setting. An analysis of the generated questions shows that only 6\% of the questions from the previous stage fall into this category, indicating that their removal does not drastically distort the original distribution as shown in Appendix~\ref{discard_distributions}.
    \item \textbf{Human Verification}: The remaining candidate QA pairs were subjected to a human verification and refinement process. Using a custom-developed annotation GUI as shown in Figure~\ref{fig:annotation_revision}, human annotators with expertise in data analysis reviewed each component. Their tasks included: (1) revising the natural language question for clarity and precision; (2) validating, debugging, and refining the Python code for correctness and efficiency; and (3) verifying the final answer derived from the code. During this stage, annotators also filtered out questions for qualitative reasons, such as leading to uninformative answers, being too ambiguous to permit a definitive answer, or requiring external knowledge that was unavailable. This human-in-the-loop process yielded a curated set of 211 high-quality QA pairs with 178 datasets. Furthermore, all the questions were rephrased by separating the output format (e.g. bar chart, list of tuples) from the question, enhancing the naturalness, and paraphrasing the column names mentioned in the questions. As a final quality control measure, a second group of annotators with much experience in data science, who were not involved in the initial revision phase, performed a concluding review by using the different annotation GUI as presented in Figure~\ref{fig:annotation_check}. This step was designed to validate the quality and logical soundness of the final QA pairs, with a particular focus on ensuring the Python code was robust and accurately addressed the corresponding question.
\end{enumerate}

After the question scoring stage, we generated a total of 1,840 candidate questions, from which we curated 211 high-quality QA pairs, requiring 9,246 LLM calls in total. A detailed breakdown of the reasons for discarding candidate questions is provided in Appendix~\ref{discard_distributions}.

\subsubsection{Annotation of Insight}
\label{annotation_insight}

Establishing a ground truth for insight generation is inherently more complex than for question answering. The subjective nature of what constitutes a meaningful finding makes achieving consensus difficult, posing a significant challenge for both automated generation and evaluation. To address this limitation, we adopted official reports that accompany the open datasets. These human-authored documents contain the key findings and conclusions originally derived by specialists. Our process involved curating six datasets (Table~\ref{tab:datasets_data_insight}) that included such reports. We then systematically extracted the principal findings from each document depending on the representation of the key findings in the report. If the report presents insights as bullet points, we directly treated each bullet point as one insight. If the report expresses findings in free text, we uploaded the report to NotebookLM~\cite{notebooklm} and prompted it to suggest ten insights from the results sections, followed by the manual verification of the quality of the extracted insights. We then synthesized the extracted insights into a standardized set of declarative sentences, and converted the set of sentences into a summary via Gemini 2.5 Flash~\cite{comanici2025gemini25pushingfrontier}. The set of insights and the summary together form the ground-truth corpus for our insight generation task.

\subsection{Benchmark Statistics}

\begin{figure*}[t!]
\centering
\includegraphics[width=1.0\linewidth]{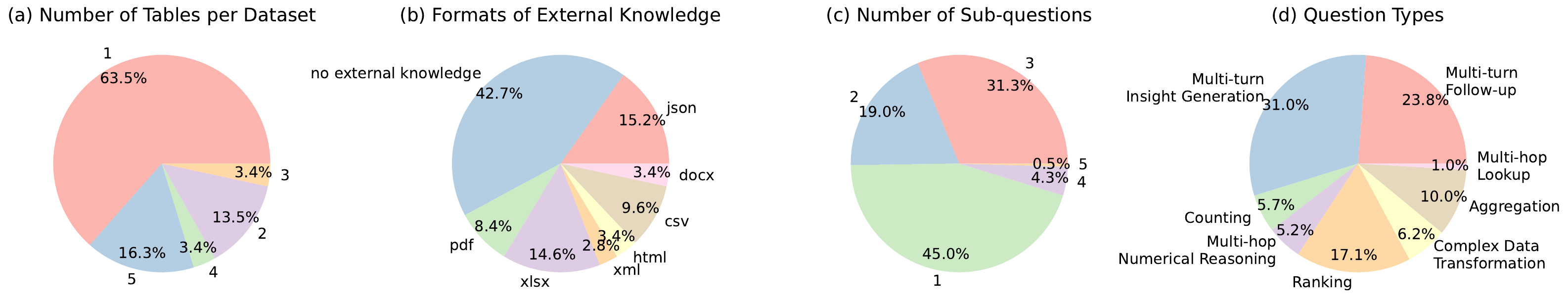}
\caption{Distributions of (a) number of tables per dataset, (b) format of external knowledge, (c) number of sub-questions, and (d) question types.}
\label{fig:distributions}
\end{figure*}

\textbf{Dataset Statistics}: A statistical overview of the datasets in DataGovBench is presented in Table~\ref{tab:statistics}. The benchmark comprises 178 unique datasets, featuring tables with an average of approximately 210K rows and 18 columns. The scale of the data is substantial, with the largest table containing up to 11.9M rows and 213 columns that exceed those found in most existing table QA benchmarks. The distribution of original open data websites is presented in Table~\ref{tab:data_source_distribution}. 
Figure~\ref{fig:distributions} (a) illustrates the distribution of tables per dataset; notably, over 36\% of the datasets are multi-tabular, with five tables as the most frequent configuration. Furthermore, as shown in Figure~\ref{fig:distributions} (b), over 57\% of the datasets are accompanied by external knowledge to aid data interpretation, provided in various formats (e.g. PDF, XLSX). These characteristics---including large, multi-table schemas and the integration of external knowledge---underscore the alignment with realistic data analysis scenarios.

\begin{wraptable}{r}{0.43\textwidth}
    \centering
    \caption{Statistics of datasets \& tasks}
    \label{tab:benchmark_stat}
    \scalebox{0.90}{\begin{tabular}{ll}
        \toprule
        \textbf{Properties} & \textbf{Value} \\
        \midrule
        \#Datasets & 178 \\
        \#Average Rows & 210K \\
        \#Max Rows & 11.9M \\
        \#Average Columns & 18.4 \\
        \#Max Columns & 213 \\
        \midrule
        \#Datasets for Table QA & 173 \\
        \#Datasets for Table Insight & 6 \\
        \#Simple Questions & 95 \\
        \#Decomposable Questions & 116 \\
        \#Individual Questions & 414 \\
        \bottomrule
    \end{tabular}
    }
\label{tab:statistics}
\end{wraptable}

\textbf{Task Statistics}: Table~\ref{tab:statistics} provides a statistical summary of tasks in the benchmark. The Table QA task includes 211 question sets, which are categorized into 95 simple questions and 116 decomposable questions. When these decomposable questions are broken down into their constituent parts, the benchmark contains a total of 414 individual questions. The distribution of sub-questions per question set is detailed in Figure~\ref{fig:distributions} (c), which shows that over 55\% of all question sets include multiple sub-questions. Figure~\ref{fig:distributions} (d) also presents the distribution of question types. For the Table Insight task, a curated subset of six datasets, each accompanied by a report from domain experts, is designated for insight generation evaluation.

\section{Experimental Setup}
\label{experimental_setup}

Our benchmark evaluates performance on two distinct tasks: \textit{Table QA} and \textit{Table Insight}. For Table QA, models receive a user question that specifies the desired output format (text or visualization). In decomposable questions, the preceding conversational history is also provided. The goal is to produce a precise textual or visual answer. The Table Insight task challenges models to generate a list of findings and a summary directly from the given files. 
Implementation details, including model configurations, hyperparameters, and settings for fair comparison are available in Appendix~\ref{implementation_details}.

\subsection{Evaluation Comparisons for Table QA}

We evaluate a range of baselines, from general-purpose LLMs to specialized agents.

\textbf{LLM}: We evaluate a diverse set of LLMs to investigate the capability of table reasoning. We selected open-source models from several categories: general-purpose (Llama 3.1~\cite{grattafiori2024llama3herdmodels}, DeepSeek-R1~\cite{deepseekai2025deepseekr1incentivizingreasoningcapability}, and Qwen3~\cite{yang2025qwen3technicalreport}), code generation-specific (Devstral~\cite{devstral} and Qwen3-Coder~\cite{qwen3coder}), and table-specific (TableGPT2~\cite{su2024tablegpt2largemultimodalmodel}). We also include high-performance closed-source models, namely GPT-4o ~\cite{openai2024gpt4ocard}, GPT-5.1~\cite{gpt5.1}, Claude Sonnet 4.6~\cite{claude4.6}, and Gemini-2.5 Flash ~\cite{comanici2025gemini25pushingfrontier}.

\begin{figure*}[t!]
\centering
\includegraphics[width=0.95\linewidth]{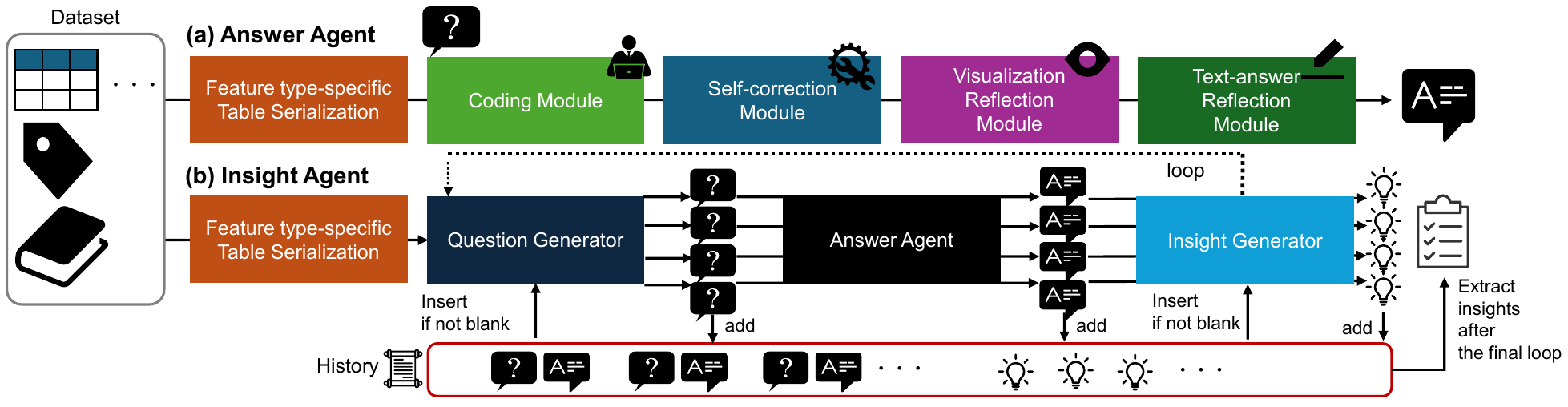}
\caption{Architecture of (a) Answer Agent and (b) Insight Agent}
\label{fig:aget_architecture}
\end{figure*}

\textbf{Answer Agent}: We leverage the Answer Agent, designed to robustly generate Python code for answering questions. The agent is composed of a practical assembly of existing following mechanisms shown in Figure~\ref{fig:aget_architecture} (a).

\begin{itemize}[leftmargin=*]
    \item Feature type-specific table serialization: This module first processes the raw tables using the table serialization based on feature types as detailed in Appendix~\ref{ft_serialization}. The resulting structured, textual representation of the data is then utilized as input for the subsequent modules.
    \item Coding: With the serialized table, metadata, external knowledge, and the input question, this module generates Python code to produce an answer. It incorporates a self-correction mechanism: if code execution fails, a subsequent LLM is called to revise the code based on the error message with up to three revision attempts allowed.
    \item Reflection: This module addresses cases where successfully executed code produces semantically incorrect outputs (e.g., a visualization with no data points or calculation resulting in NaN). A Vision-Language Model (VLM) or Multimodal Large Language Model (MLLM) is employed to analyze visual outputs, while an LLM analyzes textual results. If an issue is detected, the module triggers a revision loop to refine the code, and this can be repeated to three times.
\end{itemize}

We initially performed preliminary experiments with the specialized table agents as baselines, including InfiAgent-DABench~\cite{infida-bench} and tablegpt-agent~\cite{su2024tablegpt2largemultimodalmodel}; however, their performance on our benchmark was near-zero due to the complexity of our benchmark, so they were excluded from the final comparison.

\subsection{Evaluation Comparisons for Table Insight}

For the Table Insight task, we evaluate the following baseline agents. We employ only closed-source LLMs in these agents, given the relatively low performance of open-source models on Table QA.

\textbf{AgentPoirot}~\cite{insight-bench}: This agent is designed for goal-oriented insight generation. It operates by first extracting the data schema and then generating a set of high-level questions. For each question, it generates an answer, interprets it, and recursively poses follow-up questions to dive deeper, and finally summarizes the obtained insights. This process follows a tree-like exploration structure (Figure~\ref{fig:insight_process} (a)), where each branch represents a dive into a analytical path.

\textbf{Insight Agent}: We employ the Insight Agent shown in Figure~\ref{fig:aget_architecture} (b) that extends the core ideas of AgentPoirot in a applicable way. Similar to AgentPoirot, it iteratively generates questions, produces answers, and derives insights. The agent begins by generating the fixed number of high-level questions, which are then processed by our Answer Agent to obtain correct answers. Insights are subsequently synthesized from multiple QA pairs. These initial insights then seed the generation of new follow-up questions by combining multiple insights, continuing the cycle, ending by the summarization. Unlike AgentPoirot's tree-structured approach, the Insight Agent employs a directed acyclic graph (DAG)-based approach as explained in Figure~\ref{fig:insight_process} (b), as the generation of new questions and insights selects and aggregates the context from all previously generated information instead of single insight or QA pair in the previous depth. 

\subsection{Evaluation Metrics}

To assess the performance of agents and LLMs on our benchmark, we compare their outputs against the ground-truth references. Distinct evaluation protocols are employed for Table QA and Table Insight tasks. 

\textbf{Table QA}: The QA task is evaluated on accuracy under two settings: \textit{Whole}, where all sub-questions in a decomposable question must be correct, and \textit{Individual}, which measures sub-question-level accuracy.
Correctness is determined by the modality of the answer. Text-based answers are judged by Exact Match (EM). Visualizations are evaluated using an MLLM-as-a-judge protocol, where four MLLMs (GPT-4o, GPT-4o-mini, Gemini 2.5 Flash, and Gemini 2.5 Pro~\cite{comanici2025gemini25pushingfrontier}) assess the semantic equivalence between the predicted and ground-truth outputs. The judges are provided both the rendered images and their source code, and a prediction is deemed correct upon a majority consensus, requiring positive assessments from at least three of the four models.

\textbf{Table Insight}: To evaluate the quality of generated insights, we adopt the methodology from InsightBench~\cite{insight-bench} computing LLaMA-3-Eval scores. We employ GPT-4o as the evaluator by replacing LLaMA3-70b~\cite{grattafiori2024llama3herdmodels}. We verified that the evaluator aligns with human perception through a meta-evaluation, as shown in Appendix~\ref{meta_evaluation}. The evaluation is conducted at two levels of granularity:

\begin{itemize}[leftmargin=*]
    \item Summary-level Score: This metric assesses the holistic quality of the entire set of generated insights by comparing it against the ground-truth summary.
    \item Insight-level Score: This metric provides a more fine-grained analysis. It measures the semantic alignment between each individual ground-truth insight and the most relevant insight from the predicted set, with the final score being the average of these individual comparisons.
\end{itemize}

\subsection{Implementation Details}
Answer Agent requires a VLM or a MLLM within its Reflection Module to validate visual outputs. For experiments involving closed-source models, we utilized their native multimodal capabilities across all modules. For the open-source agent configurations, we paired various LLMs with a specialized VLM, Chart-R1-7B~\cite{chen2025chartr1}.
For the Table Insight task, each experiment was executed five times per model, and the scores were averaged across these runs to ensure the stability of our result. All results were obtained with the model temperature set to 0.0 to promote deterministic output. The other details about LLM configurations and hyperparameters are available in Appendix~\ref{implementation_details}.

\section{Evaluation Results}

\begin{table}[t!]
\centering
\scalebox{0.70}{
\begin{tabular}{l|cccc|cccc}
\toprule
\textbf{LLM} & \multicolumn{4}{|c|}{\textbf{Table QA}} & \multicolumn{4}{|c}{\textbf{Table Insight}} \\
& \multicolumn{2}{c}{w/o Answer Agent} & \multicolumn{2}{c|}{w/ Answer Agent} & \multicolumn{2}{c}{AgentPoirot} & \multicolumn{2}{c}{Insight Agent} \\
& Whole & Individual & Whole & Individual & Insight & Summary & Insight & Summary \\
\midrule
\multicolumn{9}{c}{\textbf{Closed-source LLMs}} \\
\midrule
Gemini 2.5 Flash & 0.310 & 0.401 & \textbf{0.393} & 0.502 & 0.283 & 0.359 & 0.315 & 0.405 \\
Claude Sonnet 4.6 & 0.337 & 0.478 & 0.389 & \textbf{0.534} & \textbf{0.338} & 0.397 & 0.323 & \textbf{0.453} \\
GPT-5.1 & 0.289 & 0.411 & 0.348 & 0.489 & 0.295 & 0.355 & 0.311 & 0.401 \\
GPT-4o & 0.242 & 0.333 & 0.270 & 0.391 & 0.292 & 0.345 & 0.319 & 0.399 \\
\midrule
\multicolumn{9}{c}{\textbf{Open-source LLMs}} \\
\midrule
Qwen3-30B & 0.134 & 0.216 & 0.199 & 0.309 & \multicolumn{4}{c}{\multirow{5}{*}{\bigslashcell[5.3cm]}} \\
Qwen3-Coder-30B & 0.123 & 0.191 & 0.186 & 0.280 & & & & \\
Devstral-Small & 0.152 & 0.221 & 0.186 & 0.316 & & & & \\
DeepSeek-R1-14B & 0.038 & 0.061 & 0.095 & 0.140 & & & & \\
Llama3.1-8B & 0.019 & 0.017 & 0.019 & 0.034 & & & & \\
TableGPT2-7B & 0.057 & 0.088 & 0.066 & 0.109 & & & & \\
\bottomrule
\end{tabular}
}
\caption{Main results of Table QA and Table Insight with LLMs and specific agents. \textit{w/o Answer Agent} is without the agentic support. }
\label{tab:main_result}
\end{table}

\subsection{Quantitative Evaluation Results}
The main results for the Table QA and Table Insight tasks are presented in Table~\ref{tab:main_result}, where \textit{w/o Answer Agent} means that the LLM is executed to generate the Python codes once based on the first 10 rows of the tables instead of the specific serialization.
For the Table QA task, LLMs with the agentic support by Answer Agent consistently outperforms the base LLMs. With Gemini 2.5 Flash, for instance, it achieves relative improvements of approximately 27\% in the Whole setting and 25\% in the Individual setting. This suggests that a structured agentic framework is crucial, as standalone LLM reasoning is insufficient for such complex tasks. Despite these gains, the top absolute score in the Whole setting remains below 0.4, highlighting the difficulty of the benchmark. Among the open-source models, Qwen3-30B achieves the highest score in the Whole setting, while Devstral-small performs best in the Individual setting. However, their performance still lags behind that of the closed-source models. Notably, TableGPT2-7B, a model specialized for tabular data, scores below 0.1 even when paired with our Answer Agent.

In the Table Insight task, Claude Sonnet 4.6 has the best contributions to the agentic approach, achieving 0.338 for the Insight-level score by AgentPoirot and 0.453 for the Summary-level score by Insight Agent. However, the absolute scores remain low even with the top-performing model (below 0.35 for the Insight-level score and below 0.5 for the Summary-level score), indicating substantial challenges remain in automated insight generation.

\subsection{Qualitative Analysis}

\subsubsection{Error Analysis of Table QA}
\label{tableqa_analysis}

\begin{wrapfigure}{r}{0.50\textwidth}
\centering
\includegraphics[width=\linewidth]{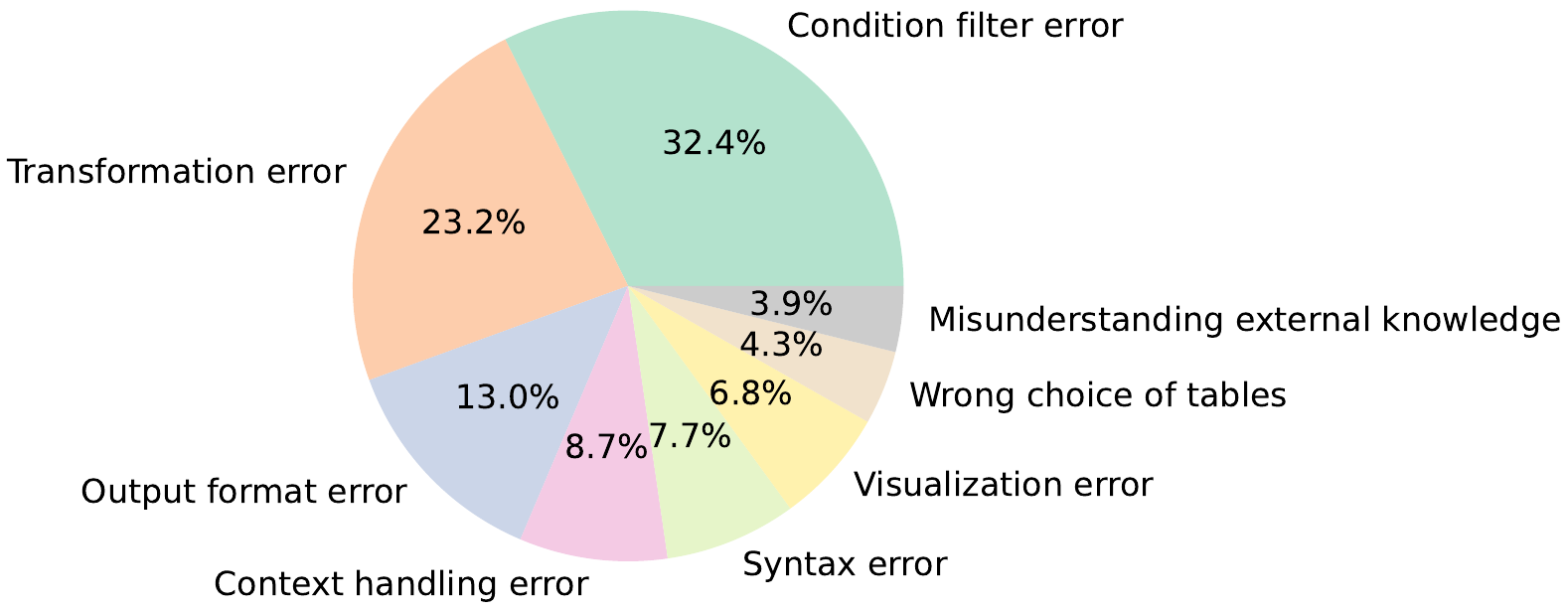}
\caption{Error distribution with Answer Agent}
\label{fig:error_analysis}
\end{wrapfigure}

We conducted an error analysis on the incorrect answers from Gemini 2.5 Flash with Answer Agent, and the results are categorized in Figure~\ref{fig:error_analysis}.

The most prevalent issue, \textbf{Condition Filter Error (32.4\%)}, occurs when the model fails to apply implicit conditions not explicitly stated in the question. A common example involves datasets with aggregated and disaggregated data (e.g., population counts for `male', `female', and `total'); models often fail to apply appropriate filters to avoid double-counting, leading to incorrect calculations. 
The second most frequent category is \textbf{Transformation Error (23.2\%)}, which involves failures in data wrangling. Common mistakes include parsing-related failures various datetime formats (e.g. day-first format) by using \verb|pandas.to_datetime| method or neglecting to convert numerical strings (e.g., "1,234,567") into integer or float types. To mitigate these errors, the more comprehensive yet efficient view of tables (e.g. exploratory data analysis results) is required in addition to the feature type-based representations for future work. 

Errors also arise from the inherent complexity of the tasks. \textbf{Context Handling Errors (8.7\%)} occur in decomposable questions where the model incorrectly uses the output from a flawed previous turn though the generated logic is correct in most cases. \textbf{Visualization Errors (6.8\%)} typically involve incorrect axis ranges, such as a timeline that does not match the period specified in the question.
Finally, the complexity of the benchmark's data structure leads to specific errors. These include \textbf{Wrong Choice of Tables (4.3\%)} in multi-table scenarios. This error occurs when the generated code fails to select the correct table from a multi-table dataset based on information provided in the metadata. For example, a dataset may contain separate tables for annual statistics, with the year covered by each table specified only in the metadata. An error arises if a question pertains to a specific year, but the model fails to refer to the metadata and consequently queries the wrong table.
\textbf{Misunderstanding External Knowledge (3.9\%)}, where the model fails to correctly apply information from provided data dictionaries to interpret the data.

\subsubsection{Fine-grained Analysis of Table Insight}
\label{tableinsight_analysis}

\begin{figure*}[t!]
\centering
\includegraphics[width=1.0\linewidth]{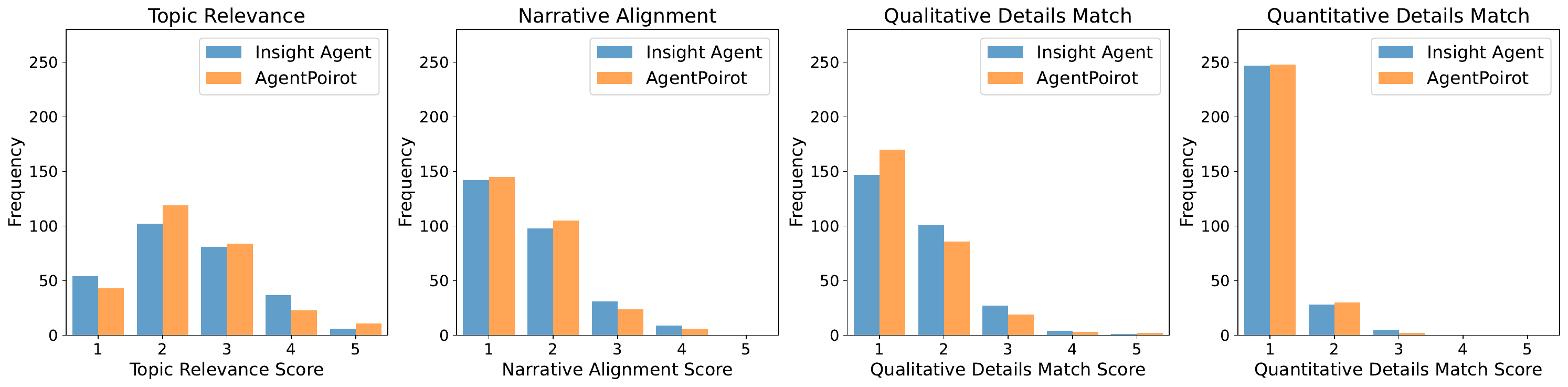}
\caption{Insight perspective distributions by Insight Agent and AgentPoirot.}
\label{fig:insight_distribution}
\end{figure*}

While the metrics provides a single value to measure the alignment between predicted and target insights, we conduct a more fine-grained analysis to understand model capabilities. We decompose the evaluation into four distinct perspectives:
\textbf{Topic Relevance} (\textit{Does the predicted insight address the same topic as the target?}), \textbf{Narrative Alignment} (\textit{Does the prediction make the same core argument or conclusion as the target?}), \textbf{Qualitative Details Match} (\textit{Does the prediction mention the same specific names or entities as the target?}), and \textbf{Quantitative Details Match} (\textit{Does the prediction mention the same specific quantitative values as the target?}).
For each of the 280 pairs that are randomly sampled~\footnote{$5$ executions $\times$ $56$ GT insights (10 per 5 datasets, 6 for 1 dataset)}, we prompted GPT-4o to score the prediction on each perspective using a 1–5 scale. The score distributions for both Insight Agent and AgentPoirot are presented in Figure~\ref{fig:insight_distribution}.

The results show that both agents perform relatively well on Topic Relevance, the highest-level criterion. However, their performance drops on Narrative Alignment and Qualitative Details Match, where high scores (3–5) are less frequent. This suggests that although the agents identify the general topic of the target insight, they frequently fail to reproduce the core argument or the qualitative entities that support it. As a result, many generated insights lack the fundamental elements needed to convey the intended conclusion, indicating that the predictions remain misaligned with the target insights. This interpretation is supported by Table~\ref{tab:a2a_comparison} where the generated insights lacking the correct core argument and qualitative entities receive lower scores than those that include them.
Finally, both agents fail on Quantitative Details Match, with none of the predicted insights scoring 3 or higher. This highlights that both agents lack the ability to accurately calculate and present specific numerical values in their generated insights.

These findings reveal two missing capabilities of LLM agents when handling real-world tables: narrative-level reasoning and accurate fact retrieval. For the former, both agents struggle to identify the central analytical argument underlying the target insight, suggesting that stronger reasoning over complicated tabular data is required. The latter capability is closely related to Table QA; therefore, the same observations discussed in the previous subsection apply here. In particular, the results suggest that a more comprehensive yet compact representation of tables is necessary to reduce various types of errors.

\subsection{Ablation Study}

\subsubsection{Components of Answer Agent} 
\begin{wraptable}{r}{0.70\textwidth}
    \centering
    \caption{Ablation Study of Answer Agent}
    \label{tab:serialization}
    \scalebox{0.60}{\begin{tabular}{l|cc}
        \toprule
        \textbf{Settings} & \textbf{Gemini 2.5 Flash} & \textbf{GPT-4o} \\
        \midrule
        Baseline & 0.310 & 0.242 \\
        + Schema & 0.308 & 0.244 \\
        + Schema + More number of rows & 0.275 & 0.232 \\
        + Serialization & 0.360 & 0.257 \\
        + Serialization + Reflection & 0.379 & 0.267 \\
        + Serialization + Reflection + Self-correction (Answer Agent) & 0.393 & 0.270 \\
    \bottomrule
    \end{tabular}
    }
\label{tab:effectiveness_serialization}
\end{wraptable}
To assess the contribution of each Answer Agent module quantitatively, we conducted an ablation study against a naive baseline using the first 10 table rows (Table~\ref{tab:effectiveness_serialization}) for the Python code generation.
Adding the table schema consisting of column names, data types, and statistics on missing and unique values does not affect the performance. We further increased the number of provided rows whose number was increased up to a maximum of 200, adjusted as needed to fit within the LLM context window. The performance degraded noticeably, suggesting that LLMs struggle to identify the key information needed for solution construction when faced with a larger pool of numerical or textual entries, compared to the first 10 rows.
In contrast, the feature type-specific serialization led to gains over the baseline by providing a compact and reliable representation of the table that avoids the baseline’s tendency to rely on guessed values. This finding further supports the idea that a more comprehensive and efficient representation of tables leads to higher performance, as discussed in Section~\ref{tableqa_analysis}.
Adding the Reflection module further improved performance across models by enabling the system to capture implicit constraints and infer missing reasoning steps.

\subsubsection{SQL vs. Python for Table Question Answering} 

\begin{wraptable}{r}{0.35\textwidth}
    \centering
    \caption{SQL vs. Python}
    \label{tab:serialization}
    \scalebox{0.70}{\begin{tabular}{l|cc}
        \toprule
        \textbf{LLM} & \textbf{SQL} & \textbf{Python} \\
        \midrule
        Gemini 2.5 Flash & 0.241 & 0.545 \\
        GPT-4o & 0.145 & 0.303 \\
    \bottomrule
    \end{tabular}
    }
\label{tab:python_vs_sql}
\end{wraptable}

Text-to-SQL has been a primary approach for tackling table question answering, where natural language questions are translated into SQL queries. SQL is widely used for efficient data manipulation and retrieval, making it meaningful to investigate whether SQL is a more suitable output format than Python code for LLM-based table reasoning.
In this study, we exclude visualization-based QA pairs from DataGovBench, as SQL does not provide native functions for generating visualizations. As a result, the evaluation is conducted on 145 text-based QA pairs. The LLM prompt includes feature type-specific serialization, as information about feature types and data types helps the model construct proper SQL queries.
As shown in Table~\ref{tab:python_vs_sql}, the results indicate that SQL performs significantly worse than Python code. Error analysis of the generated SQL queries reveals that syntax errors and incorrect column selections occur more frequently than in generated Python code.


\section{Related Works}

\textbf{Benchmarks for Table Question and Answering}. Research in table-based question answering has been largely driven by a series of influential benchmarks. Early work such as WikiTableQuestions~\cite{wtq} established the task of answering natural language questions over Wikipedia tables. While foundational, this benchmark is limited to single tables and relatively simple questions. Although benchmarks like FeTaQA~\cite{nan-etal-2022-fetaqa} and OTT-QA~\cite{ottqa} introduced tasks requiring more complicated reasoning, they still focus on small-scale Wikipedia tables.
The subsequent development of text-to-SQL benchmarks marked a significant leap in complexity. Spider~\cite{yu-etal-2018-spider} and BIRD~\cite{bird} became the standards requiring models to generate complex SQL queries. More recently, developments in LLMs have enabled models to generate coherent Python code, leading to the construction of benchmarks that assess data analysis capabilities~\cite{wu2025tablebench,infida-bench,oses-grijalba-etal-2024-question}.
While these benchmarks were instrumental in advancing table reasoning capabilities, they do not comprehensively capture the challenges of real-world data analysis. The datasets are typically well-structured in scale. They often lack practical characteristics, such as rich metadata and supplementary external knowledge. Furthermore, many do not address the large multi-tabular datasets. DataGovBench comprehensively targets these limitations by incorporating all of these features: large tables, multi-table schemas, metadata, and extenral knowledge.

\textbf{Benchmarks for Table Insight Generation}. Automated insight generation is a nascent and challenging area to benchmark, as the subjective nature of an "insight" complicates objective evaluation~\cite{data2dashboard,majumder2024discoverybenchdatadrivendiscoverylarge}. Recent work like InsightBench~\cite{insight-bench} and T2R-Bench~\cite{zhang-etal-2025-t2r} have advanced this area by introducing benchmarks for table insight generation. However, they do not fully capture the complexities of real-world data analysis. InsightBench focuses primarily on small-scale tables, while T2R-Bench lacks supporting metadata and external knowledge that are often required to interpret tabular data. In contrast, DataGovBench provides a more comprehensive benchmark by incorporating large-scale collections of tables together with relevant external knowledge sources.

\section{Conclusion}

To address the critical lack of realism in existing benchmarks, we introduced \textit{DataGovBench}, a new benchmark built from open data. It features large, multi-tabular datasets, and incorporates external knowledge, and formalizes two key tasks: complex Question Answering (with decomposable questions and visualizations) and a novel Insight Generation task grounded in reports by domain specialists. Our extensive evaluation reveals that even state-of-the-art models with and without the agentic support struggle with low QA accuracy, highlighting the difficulity of both tasks. A detailed qualitative analysis provides a clear roadmap for future research. Future efforts should focus on both addressing these issues and efficiently scaling the benchmark's size to ensure more robust evaluations. We believe DataGovBench will serve as a catalyst steering research toward building more robust agents capable of handling real-world data complexities.

\subsubsection{Use of Large Language Models} We used LLM for the grammar correction and the words refinement to
enhance the quality of the paper.

%
%
%

\bibliographystyle{splncs04}
\bibliography{reference}

%

\appendix

\section{Benchmark Construction}
\label{app_benchmark_construction}

\subsection{Data Curation}
\subsubsection{Data Collection}
Table~\ref{tab:websites} lists 53 websites, from which we downloaded all datasets along with their metadata via APIs (e.g., CKAN API~\footnote{\url{https://github.com/ckan/ckanapi}}).

\subsubsection{Data Filtering}
\label{sec:data_filtering}
All downloaded datasets were subjected to a rigorous filtering and curation process to select those most suitable for real-world data analytics tasks. This process involved three main stages: dataset filtering, external knowledge identification, and metadata standardization. First, data filtering is conducted based on the following conditions:

\begin{itemize}[leftmargin=*]
    \item The dataset had to contain at least one CSV file that was correctly formatted and readable by the \verb|pandas.read_csv| method. Files that were HTML or XML in content despite having a .csv suffix were excluded.
    \item At least one CSV file within the dataset was required to have more than 5,000 rows. Additionally, tables with five or more blank columns were discarded.
    \item Each dataset needed to be accompanied by a textual description. The license was also required to permit redistribution; for datasets from Data.gov where the license was often unspecified in the metadata, we performed manual verification on the source webpage.
    \item ArcGIS-based datasets, which are primarily geospatial, were excluded from our analysis.
\end{itemize}

Following the filtering stage, we systematically searched for external knowledge (e.g., data dictionaries) within each dataset using a set of heuristic rules:

\begin{itemize}[leftmargin=*]
    \item First, an automated search was performed for files with names containing "data dictionary" or "datadictionary".
    \item Next, a platform-specific rule was applied for Data.gov datasets. We observed that when a JSON file is provided alongside CSV, XML, and RDF files, it often contains column-level descriptions. In such cases, the JSON file was designated as external knowledge.
    \item If these automated heuristics failed, we performed a manual inspection of the dataset's contents to locate any other supplementary documentation that could serve as a data dictionary.
\end{itemize}

Finally, the original metadata for each curated dataset was processed and standardized. This step created a concise metadata format specific to our benchmark by removing redundant or irrelevant information from the source. The following is an example of the specific metadata from \textit{Indiana Arrest Data} of \textit{Indiana Data Hub}. 

\begin{lstlisting}[title={Converted Metadata}]
"identifier": "d39f6598-efbb-40a7-a694-6a9b8d2dc2dc"
"dataset_title": "INDIANA ARREST DATA"
"dataset_description": "This dataset is the underlying data of the Indiana Arrests Dashboard which displays counts of individuals arrested, arrests, charges by offense category, dispositions, country and time period in Indiana beginning in 2008 through the present year. \r\n\r\nArrest data comes from the Criminal History Repository System (CHRIS). Data feeding into the CHRIS system comes from three main sources. Arrest data comes from the LiveScan system, which is used for fingerprinting and capturing other pertinent information at the time of the arrest. Criminal disposition data are maintained by prosecutors in ProsLink system, and by the courts in the Odyssey system. \r\n\r\nData Notes:\r\n\r\n1. Arrest data are sent to ISP soon after the arrest occurs, but disposition data have a lag of approximately seven months as the case makes its way through the legal system. \r\n\r\n2. Text descriptions of the original offenses are provided by the arresting officer when the offender is arrested. Later, the prosecutor's office or court provides a text description of the filed offense, along with the Indiana Code title, article, chapter, and section (e.g.35-48-4-6). The filed offense may be amended later. \r\n\r\n3. Arrest County is determined by the location of the booking agency. If the booking agency is missing, then the arresting agency is used. \r\n\r\n4. The count of individuals/arrests/charges by offense category can add up to more than the grand total because one individual/arrest/charge can fall into multiple categories (e.g. DUI is counted in the \"Drug\" and \"Traffic\" categories. \r\n\r\n5. Arrest categories and subcategories are determined based on keywords found in a free text description of the offense. About 7% of offenses have a description that has not yet been categorized."
"publisher": "Indiana State Police"
"landingPage": "Indiana State Police"
"license": "Creative Commons Attribution"
"distribution": [{"file_name": "data9.csv",
                  "file_title": "ARREST DATA 2022 Q3",
                  "file_description": null,
                  "downloadURL": "https://hub.mph.in.gov/dataset/d39f6598-efbb-40a7-a694-6a9b8d2dc2dc/resource/00cd698d-e26b-458a-861b-4c355b77ab20/download/isp_arrest_data_2022_q3.csv",
                  "accessURL": "https://hub.mph.in.gov/dataset/d39f6598-efbb-40a7-a694-6a9b8d2dc2dc/resource/00cd698d-e26b-458a-861b-4c355b77ab20/download/isp_arrest_data_2022_q3.csv"},
                 {"file_name": "data37.csv",
                  "file_title": "ARREST DATA 2015 Q3",
                  "file_description": null,
                  "downloadURL": "https://hub.mph.in.gov/dataset/d39f6598-efbb-40a7-a694-6a9b8d2dc2dc/resource/8b2b54fe-363a-46f7-9c3b-197cce01616f/download/isp_arrest_data_2015_q3.csv",
                  "accessURL": "https://hub.mph.in.gov/dataset/d39f6598-efbb-40a7-a694-6a9b8d2dc2dc/resource/8b2b54fe-363a-46f7-9c3b-197cce01616f/download/isp_arrest_data_2015_q3.csv"},
                 {"file_name": "data20.csv",
                  "file_title": "ARREST DATA 2019 Q4",
                  "file_description": null,
                  "downloadURL": "https://hub.mph.in.gov/dataset/d39f6598-efbb-40a7-a694-6a9b8d2dc2dc/resource/bd011a33-0652-4ad7-8d90-6c1019d6385c/download/isp_arrest_data_2019_q4.csv",
                  "accessURL": "https://hub.mph.in.gov/dataset/d39f6598-efbb-40a7-a694-6a9b8d2dc2dc/resource/bd011a33-0652-4ad7-8d90-6c1019d6385c/download/isp_arrest_data_2019_q4.csv"},
                 {"file_name": "data15.csv",
                  "file_title": "ARREST DATA 2021 Q1",
                  "file_description": null,
                  "downloadURL": "https://hub.mph.in.gov/dataset/d39f6598-efbb-40a7-a694-6a9b8d2dc2dc/resource/9c7960c6-417b-45e6-9ace-b75958dd91de/download/isp_arrest_data_2021_q1.csv",
                  "accessURL": "https://hub.mph.in.gov/dataset/d39f6598-efbb-40a7-a694-6a9b8d2dc2dc/resource/9c7960c6-417b-45e6-9ace-b75958dd91de/download/isp_arrest_data_2021_q1.csv"},
                 {"file_name": "data14.csv",
                  "file_title": "ARREST DATA 2021 Q2",
                  "file_description": null,
                  "downloadURL": "https://hub.mph.in.gov/dataset/d39f6598-efbb-40a7-a694-6a9b8d2dc2dc/resource/1ff2cf5f-69ef-4139-bcb4-036f66787172/download/isp_arrest_data_2021_q2.csv",
                  "accessURL": "https://hub.mph.in.gov/dataset/d39f6598-efbb-40a7-a694-6a9b8d2dc2dc/resource/1ff2cf5f-69ef-4139-bcb4-036f66787172/download/isp_arrest_data_2021_q2.csv"}],
"external_knowledge": ["data68.xlsx"]}
\end{lstlisting}

\begin{table}
    \centering
    \caption{List of Open Data Websites}
    \begin{adjustbox}{max width=1.0\textwidth}
    \begin{tabular}{lp{0.55\linewidth}}
        \toprule
        \textbf{Websites} & \textbf{URL} \\
        \midrule
        Data.gov & \small{\url{https://data.gov/}} \\
        California Open Data Portal & \small{\url{https://data.ca.gov/}} \\
        Hawaii Open Data & \small{\url{https://opendata.hawaii.gov/}} \\
        Analyze Boston & \small{\url{https://data.boston.gov/}} \\
        City of Houston Open Data & \small{\url{https://data.houstontx.gov/}} \\
        The Indiana Data Hub & \small{\url{https://hub.mph.in.gov/}} \\
        Milwaukee Open Data & \small{\url{https://data.milwaukee.gov/}} \\
        Open Data SA & \small{\url{https://data.sanantonio.gov/}} \\
        Pompano Beach Open Data & \small{\url{https://data.pompanobeachfl.gov/}} \\
        America's Education data & \small{\url{https://data.ed.gov/}} \\
        Energy Data eXchange & \small{\url{https://edx.netl.doe.gov/}} \\
        California Health and Human Services Open Data Portal & \small{\url{https://data.chhs.ca.gov/}} \\
        California Natural Resources Agency Open Data & \small{\url{https://data.cnra.ca.gov/}} \\
        U.S. Small Business Administration Open Data & \small{\url{https://data.sba.gov/}} \\
        Ireland's Open Data Portal & \small{\url{https://data.gov.ie/}} \\
        Dublinked: Open Data for the Dublin Region & \small{\url{https://data.smartdublin.ie/}} \\
        Tusla Data Catalogue & \small{\url{https://datacatalog.tusla.ie/}} \\
        DAFM Data Portal & \small{\url{https://opendata.agriculture.gov.ie/}} \\
        Central Bank of Ireland's Open Data Portal & \small{\url{https://opendata.centralbank.ie/}} \\
        Data.gov.au & \small{\url{https://data.gov.au/}} \\
        The Central Resource for SEED in NSW & \small{\url{https://www.seed.nsw.gov.au/}} \\
        Data.NSW & \small{\url{https://data.nsw.gov.au/}} \\
        NTG Open Data Portal & \small{\url{https://data.nt.gov.au/}} \\
        Data.SA & \small{\url{https://data.sa.gov.au/}} \\
        Ballarat Open Data & \small{\url{https://ballaratopendata.org.au/}} \\
        DATA VIC & \small{\url{https://www.data.vic.gov.au/}} \\
        Data WA & \small{\url{https://www.data.wa.gov.au/}} \\
        Queensland Government Publications Portal & \small{\url{https://www.publications.qld.gov.au/}} \\
        Transport Open Data & \small{\url{https://opendata.transport.nsw.gov.au/}} \\
        openAFRICA & \small{\url{https://open.africa/}} \\
        Data.gov.hk & \small{\url{https://data.gov.hk/en/}} \\
        Data.gov.uk & \small{\url{https://www.data.gov.uk/}} \\
        UK Data Service & \small{\url{https://statistics.ukdataservice.ac.uk/}} \\
        London Datastore & \small{\url{https://data.london.gov.uk/}} \\
        Open Data NI & \small{\url{https://admin.opendatani.gov.uk/}} \\
        ENTSO-E & \small{\url{https://docs.entsoe.eu/}} \\
        Journal Data Archive & \small{\url{https://journaldata.zbw.eu/}} \\
        Data.openstate.eu & \small{\url{https://data.openstate.eu/}} \\
        OPERANDUM & \small{\url{https://data-catalogue.operandum-project.eu/}} \\
        Dataportal.ponderful.eu & \small{\url{https://dataportal.ponderful.eu/}} \\
        OpenCity & \small{\url{https://opencity.in/}} \\
        New Zealand's Biological Heritage Data Repository & \small{\url{https://data.bioheritage.nz/}} \\
        Datastore.landcareresearch.co.nz & \small{\url{https://datastore.landcareresearch.co.nz/}} \\
        Open.canada & \small{\url{https://search.open.canada.ca/opendata/}} \\
        Open Govermental Portal in Alberta & \small{\url{https://www.alberta.ca/open-government-program}} \\
        Data.gov.bc.ca & \small{\url{https://catalogue.data.gov.bc.ca/}} \\
        Niagara's Open Data Catalogue & \small{\url{https://niagaraopendata.ca/}} \\
        Ontario Data Catalogue & \small{\url{https://data.ontario.ca/}} \\
        Données Québec & \small{\url{https://www.donneesquebec.ca/}} \\
        Surrey's Open Data & \small{\url{https://data.surrey.ca/}} \\
        City of Toronto's Open Data Portal & \small{\url{https://open.toronto.ca/}} \\
        Data.sustain.ubc.ca & \small{\url{https://data.sustain.ubc.ca/}} \\
        Columbia Basin Water Hub & \small{\url{https://data.cbwaterhub.ca}} \\
        \bottomrule
    \end{tabular}
    \end{adjustbox}
\label{tab:websites}
\end{table}

\subsection{Data Annotation}
\label{sec:data_annotation}

\subsubsection{Question Types}
\label{question_types}

During question generation, specific question types are provided to the LLM to guide the formulation of questions. We use the following eight question types, with each type’s name and description supplied to the LLM. \textit{Multi-turn Follow-up} and \textit{Multi-turn Insight Generation} correspond to decomposable questions.

\begin{itemize}[leftmargin=*]
    \item \textbf{Aggregation}: Questions involving aggregated answers based on the statistical operations, such as mean, sum or mode
    \item \textbf{Ranking}: Questions involving answers based on the ranking
    \item \textbf{Counting}: Questions involving answers based on counting something
    \item \textbf{Multi-hop Lookup}: Question involving extracting single cell value from the table based on multiple reasoning steps
    \item \textbf{Multi-hop Numerical Reasoning}: Questions involving numerical answers based on multiple reasoning steps
    \item \textbf{Complex Data Transformation}: Question involving complex data transformation, such as aggregation or filtering across multiple dimensions, creating new columns, filtering with context-dependent logic, resolving entity references across rows, or merging multiple tables
    \item \textbf{Multi-turn Follow-up}: Question involving multi-turn follow-up questions that build on previous answers or context from table data, requiring the model to maintain state and context across multiple interactions
    \item \textbf{Multi-turn Insight Generation}: Question involving multi-turn insight generation that requires the model to generate insights or summaries based on previous answers or context from table data, requiring the model to maintain state and context across multiple interactions. Questions in the intermediate turn ask to provide not only text-based answer but also text-based complicated statistical information (e.g. correlation) and visualization
\end{itemize}

\subsubsection{Feature type-Specific Table Serialization}
\label{ft_serialization}

Our serialization process generates a compact textual representation of a table by summarizing its global properties and providing detailed, feature type-aware information for each column. The serialized text begins with the dimensions of the tables (number of rows and columns), followed by a per-column breakdown. For each column, the serialization includes: the inferred feature type, the Pandas data type~\cite{mckinney-proc-scipy-2010}, the percentage of NaN values, and a feature-specific textual summary.
The primary feature type is determined by a Feature Type Inference (FTI). Our FTI module follows a design similar to that in~\cite{autodw}. This model classifies each column into one of 11 types: \textit{Numerical}, \textit{Categorical}, \textit{Datetime}, \textit{Sentence}, \textit{URL}, \textit{Embedded Number}, \textit{List}, \textit{Ignorable ID}, \textit{Numbers with Unit}, \textit{Numbers with Sign}, \textit{Range of Numbers}, or \textit{Formatted ID}.
The Pandas data type is inferred using the built-in \verb|pandas.api.types.infer_dtype| function. While its output would overlap with the feature types, we include it because its ability to identify mixed types (e.g., columns containing both strings and integers) serves as a key signal for potential data quality issues that would require wrangling.

The feature type-specific summary is constructed according to the inferred feature type, as follows:

\begin{itemize}[leftmargin=*]
    \item Numerical: The minimum and maximum values in the column are included.
    \item Categorical: If the column contains 20 or fewer unique categories, all are listed. Otherwise, a random sample of 20 unique categories is provided.
    \item Datetime: The earliest and latest date or time values are included.
    \item URL: No sample values are included. This is a deliberate choice to conserve context length, as full URLs are token-intensive and typically have low semantic value for general data analysis tasks.
    \item All Other Types: For all other feature types, a random sample of 10 unique values is included to provide a representative snapshot of the column's contents.
\end{itemize}

The example of the serialized text is provided in the following from \textit{E-bike Field Study} of \textit{Data.gov}.

\begin{lstlisting}[title={Text Example by Feature type-specific Table Serialization}]
1st table
Dataset title: Comma Separated Values File
Dataset description: None

Headers and values:
Number of columns: 21
Number of rows: 408363

Feature type, pandas type, ratio of missing values, and feature type-specific information is given for each column as below.

date (feature type: Datetime) (pandas type: string) (ratio of missing values: 0%): Start date is 2022-04-27 23:42:29.834000+00:00, and end date is 2022-09-23 18:31:05.502000+00:00.
lat (feature type: Numerical) (pandas type: floating) (ratio of missing values: 0%): Value range is [42.447303, 42.461437200000006].
lon (feature type: Numerical) (pandas type: floating) (ratio of missing values: 0%): Value range is [-71.3243906, -71.2562746].
spd (feature type: Numerical) (pandas type: floating) (ratio of missing values: 0%): Value range is [0.0, 23.825000000000003].
blind_turn (feature type: Categorical) (pandas type: integer) (ratio of missing values: 0%): All categories are [0, 1].
constrained_tunnel (feature type: Categorical) (pandas type: integer) (ratio of missing values: 0%): All categories are [0, 1].
narrow (feature type: Categorical) (pandas type: integer) (ratio of missing values: 0%): All categories are [0, 1].
slow_sign (feature type: Categorical) (pandas type: integer) (ratio of missing values: 0%): All categories are [0, 1].
trail_hazards (feature type: Categorical) (pandas type: integer) (ratio of missing values: 0%): All categories are [0, 1].
trail_junction (feature type: Categorical) (pandas type: integer) (ratio of missing values: 0%): All categories are [0, 1].
vehicle_conflict_point (feature type: Categorical) (pandas type: integer) (ratio of missing values: 0%): All categories are [0, 1].
walk_bike_sign (feature type: Categorical) (pandas type: integer) (ratio of missing values: 0%): All categories are [0, 1].
eb (feature type: Categorical) (pandas type: integer) (ratio of missing values: 0%): All categories are [0, 1].
uphill (feature type: Categorical) (pandas type: integer) (ratio of missing values: 0%): All categories are [0, 1].
downhill (feature type: Categorical) (pandas type: integer) (ratio of missing values: 0%): All categories are [0, 1].
passing (feature type: Categorical) (pandas type: integer) (ratio of missing values: 0%): All categories are [0, 1].
participantid (feature type: Numerical) (pandas type: integer) (ratio of missing values: 0%): Value range is [1, 37].
age (feature type: Numerical) (pandas type: integer) (ratio of missing values: 0%): Value range is [27, 65].
sex (feature type: Categorical) (pandas type: string) (ratio of missing values: 0%): All categories are [female, male].
bike_type (feature type: Categorical) (pandas type: string) (ratio of missing values: 0%): All categories are [conventional, electric].
ebike_class (feature type: Categorical) (pandas type: floating) (ratio of missing values: 56%): All categories are [1.0, 2.0, 3.0].
    
\end{lstlisting}

\subsubsection{Human Verification}
\label{human_verification}

\begin{figure*}[h!]
\centering
\includegraphics[width=0.95\linewidth]{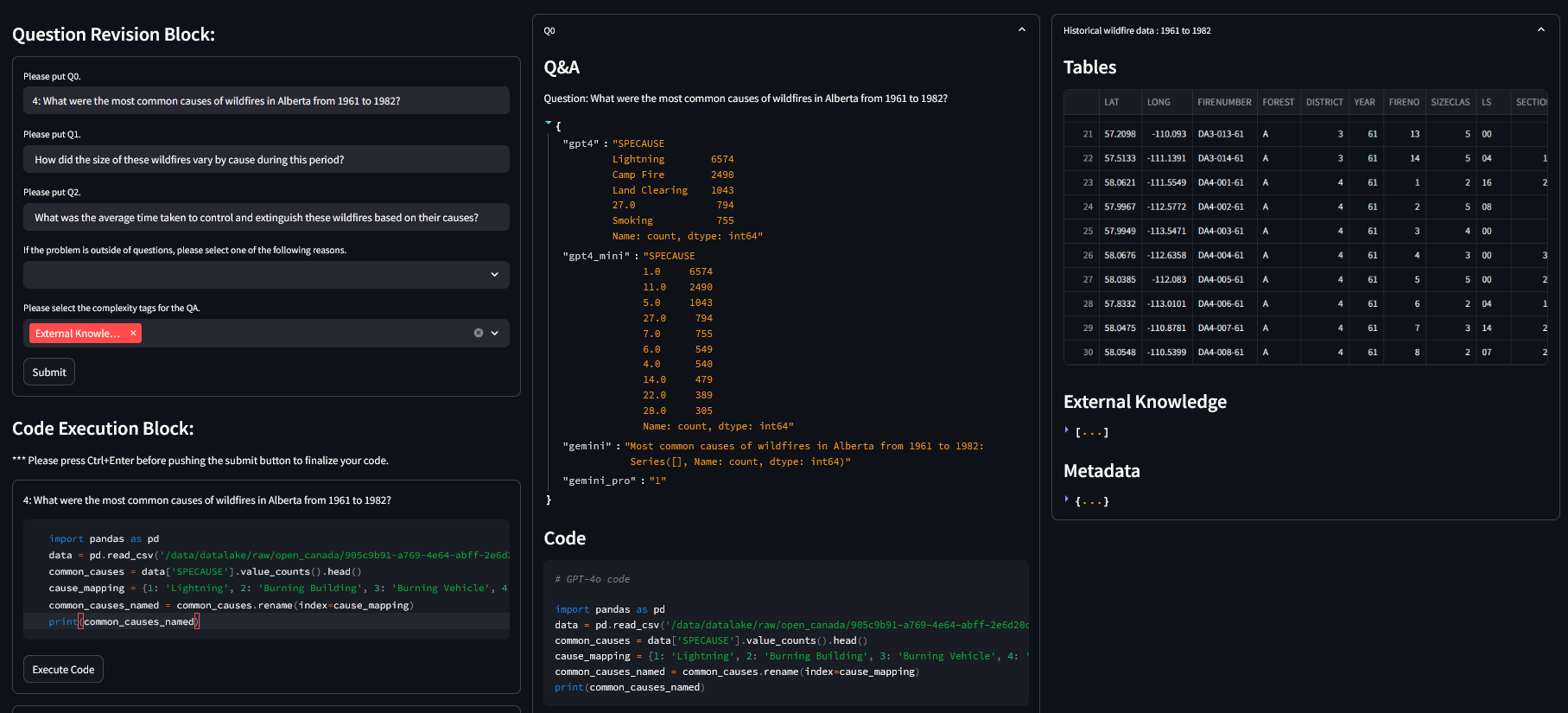}
\caption{Annotation GUI for revising questions and answers}
\label{fig:annotation_revision}
\end{figure*}

\begin{figure*}[h!]
\centering
\includegraphics[width=0.95\linewidth]{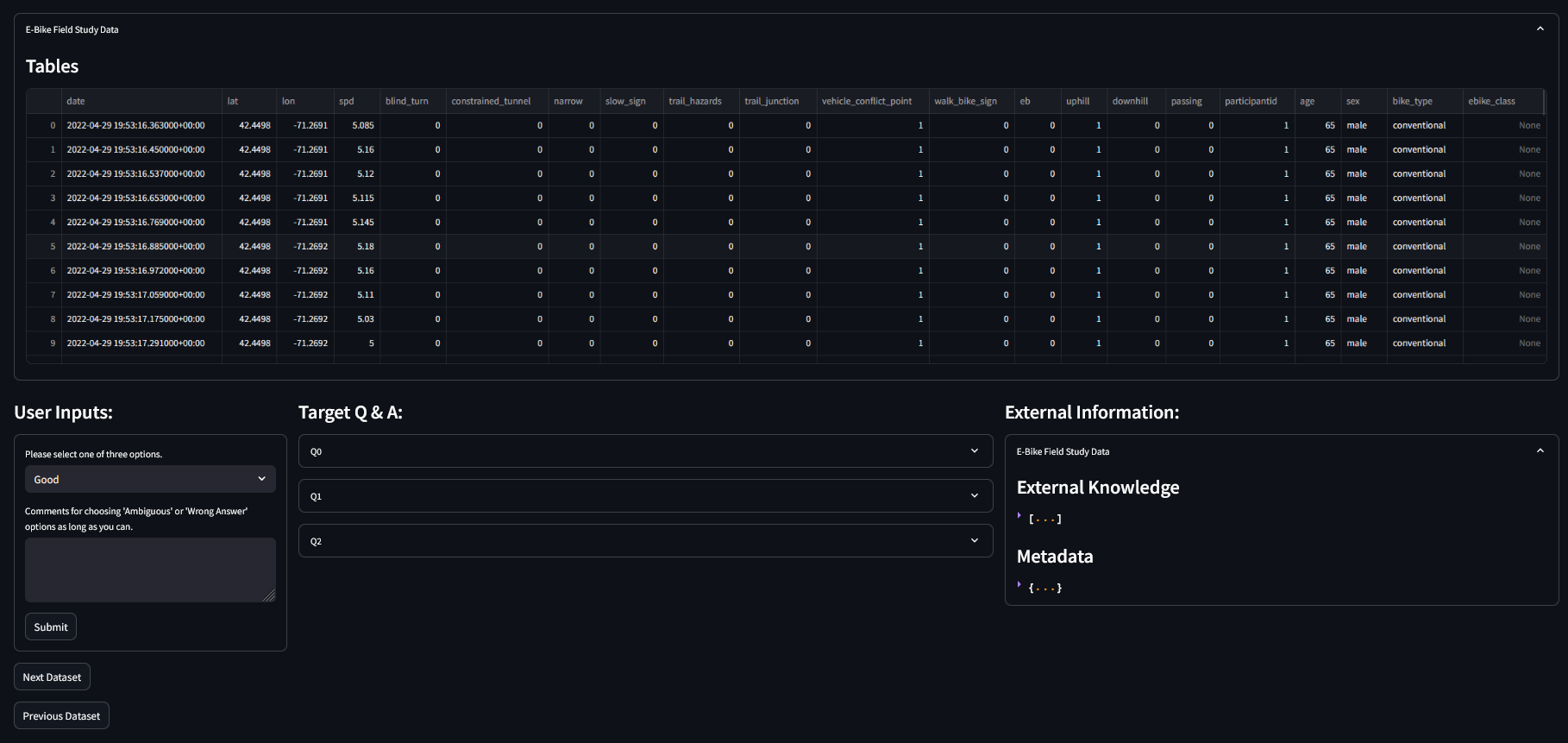}
\caption{Annotation GUI for checking the revised QA pairs}
\label{fig:annotation_check}
\end{figure*}

Figure~\ref{fig:annotation_revision} shows the annotation GUI, built with Streamlit~\footnote{\url{https://github.com/streamlit/streamlit}}, for revising questions and the Python code used to generate answers. Annotators can refer to the LLM-generated answers and code, as well as the underlying tables, metadata, and external knowledge. Figure~\ref{fig:annotation_check} presents the GUI for reviewing revised QA pairs, where annotators select one of three options—--\textit{Good}, \textit{Ambiguous}, or \textit{Wrong Answer}—--and may leave comments for the latter two. In total, we obtained 211 datasets, with their original website distribution shown in Table~\ref{tab:data_source_distribution}. Six of these datasets (Table~\ref{tab:datasets_data_insight}) are used for the Table Insight task.

\subsection{Distribution of Discarded Candidate Questions}
\label{discard_distributions}

We generated 1,840 candidate questions after the question scoring stage and curated them into 211 high-quality questions, discarding the remaining 1,629 based on the criteria below.

\begin{itemize}
    \item \textbf{Unanimous Agreement}: During the answer generation stage, we employed four LLMs to produce answers and measured consensus across models. Questions for which all LLMs produced identical answers were removed because they typically require only shallow reasoning and do not align with the intended complexity of our benchmark. Answer consensus was determined using exact string match for text-based answers and manual verification for visualization-based answers using the GUI tool shown in Figure~\ref{fig:annotation_revision}.
    \item \textbf{Insufficient External Knowledge}: Some datasets require external knowledge (e.g. data dictionary) to interpret column meanings or specific values. When such information was missing or insufficient, it became impractical to map column names and values to the generated questions, making accurate answer generation infeasible. Questions of this type were removed.
    \item \textbf{Ambiguous Question}: Questions allowing multiple plausible answers were removed to ensure benchmark clarity. For example, when a table contains both calendar-year and fiscal-year columns, a question asking for “the year” satisfying certain conditions becomes ill-posed unless the question explicitly specifies which type of year should be used.
    \item \textbf{Uninsightful Question}: We excluded questions that failed to yield analytically meaningful or practically useful insights despite being answerable. A common example occurs in geospatial datasets, where questions such as “What is the average latitude and longitude under certain conditions?” often result in a coordinate that lacks geographic or analytical relevance (e.g., a point in the ocean). Such questions were deemed non-insightful and removed.
\end{itemize}

The distribution of the reasons is shown in Table~\ref{tab:discard_distribution}.

\begin{table}
    \centering
    \caption{Distribution of Discarding Reasons}
    \begin{tabular}{ll}
        \toprule
        \textbf{Reason} & \textbf{Count and Ratio} \\
        \midrule
        Unanimous Agreement & 114 (0.062) \\
        Insufficient External Knowledge & 550 (0.299) \\
        Ambiguous Question & 354 (0.192) \\
        Uninsightful Question & 611 (0.332) \\
        \bottomrule
    \end{tabular}
\label{tab:discard_distribution}
\end{table}

\begin{table}
    \centering
    \caption{Distribution of Open Data Websites in DataGovBench}
    \begin{tabular}{ll}
        \toprule
        \textbf{Websites} & \textbf{Count} \\
        \midrule
        Open.canada & 63 \\
        Data.gov & 35 \\
        California Open Data Portal & 27 \\
        Open Govermental Portal in Alberta & 7 \\
        Data.gov.uk & 6 \\
        Analyze Boston & 5 \\
        Ontario Data Catalogue & 4 \\
        The Indiana Data Hub & 4 \\
        Data.SA & 4 \\
        Surrey's Open Data & 4 \\
        Open Data NI & 4 \\
        The Central Resource for SEED in NSW & 2 \\
        Pompano Beach Open Data & 2 \\
        Data.NSW & 2 \\
        U.S. Small Business Administration Open Data & 1 \\
        Hawaii Open Data & 1 \\
        City of Houston Open Data & 1 \\
        openAFRICA & 1 \\
        City of Toronto's Open Data Portal & 1 \\
        Milwaukee Open Data & 1 \\
        OpenCity & 1 \\
        DATA VIC & 1 \\
        Columbia Basin Water Hub & 1 \\
        \bottomrule
    \end{tabular}
\label{tab:data_source_distribution}
\end{table}

\begin{table}
    \centering
    \caption{Datasets for Table Insight}
    \begin{tabular}{p{0.55\linewidth}ll}
        \toprule
        \textbf{Dataset} & \textbf{Website} & \textbf{Domain} \\
        \midrule
        Boston Buildings Inventory & Analyze Boston & Real Estate \\
        Number of Weight Loss Surgeries Performed in California Hospital & Data.gov & Healthcare \\
        Cross-Canada Survey of Radon Concentrations in Homes & Open.canada & Environment \\
        Fixed gear sentinel fisheries program - northern Gulf of St. Lawrence & Open.canada & Marine Biology \\
        Canadian Health Measures Survey (CHMS) Human Biomonitoring Data for Environmental Chemicals & Open.canada & Environment \\
        Results from the 2023 Staffing and Non-Partisanship Survey & Open.canada & Demographics \\
        \bottomrule
    \end{tabular}
\label{tab:datasets_data_insight}
\end{table}

\section{Experimental Setup}

\subsection{Implementation Details}
\label{implementation_details}

All open-source models are sourced from the HuggingFace's \verb|transformers| library~\cite{wolf2020huggingfacestransformersstateoftheartnatural}, and experiments were conducted using 2 × 48 GB NVIDIA L40S GPUs.
Table~\ref{tab:huggingface_models} lists the API names of closed-source models and the HuggingFace model names of open-source models.

\begin{table}[h]
    \centering
    \caption{List of LLM model names in the experiments}
    \begin{tabular}{ll}
        \toprule
        \textbf{Model Name} & \textbf{API name or Huggingface model name} \\
        \midrule
        GPT-4o & \verb|gpt-4o-2024-08-06| \\
        GPT-4o-mini & \verb|gpt-4o-mini-2024-07-18| \\
        GPT-5.1 & \verb|gpt-5.1-2025-11-13| \\
        Claude Sonnet 4.6 & \verb|claude-sonnet-4-6| \\
        Gemini 2.5 Flash & \verb|gemini-2.5-flash| \\
        Gemini 2.5 Pro & \verb|gemini-2.5-pro| \\
        Devstral-Small & \verb|mistralai/Devstral-Small-2507| \\
        Qwen3-30B & \verb|Qwen/Qwen3-30B-A3B-Instruct-2507| \\
        Qwen3-Coder-30B & \verb|Qwen/Qwen3-Coder-30B-A3B-Instruct| \\
        DeepSeek-R1-14B & \verb|deepseek-ai/DeepSeek-R1-Distill-Qwen-14B| \\
        Llama3.1-8B & \verb|meta-llama/Llama-3.1-8B-Instruct| \\
        TableGPT2-7B & \verb|tablegpt/TableGPT2-7B| \\
        \bottomrule
    \end{tabular}
\label{tab:huggingface_models}
\end{table}

Insight Agent was configured to generate three initial high-level questions and perform four iterations of its question-answer-insight cycle, resulting in 12 insights. Also, the AgentPoirot baseline was configured with parameters that also yielded 12 insights. Furthermore, the summarizing LLM in Insight Agent is instructed to generate the same number of tokens as the summarized sentences from AgentPoirot for a fair comparison.

\subsection{Insight Agent}
\label{insight_agent}

\begin{figure}[t!]
\centering
\includegraphics[width=\linewidth]{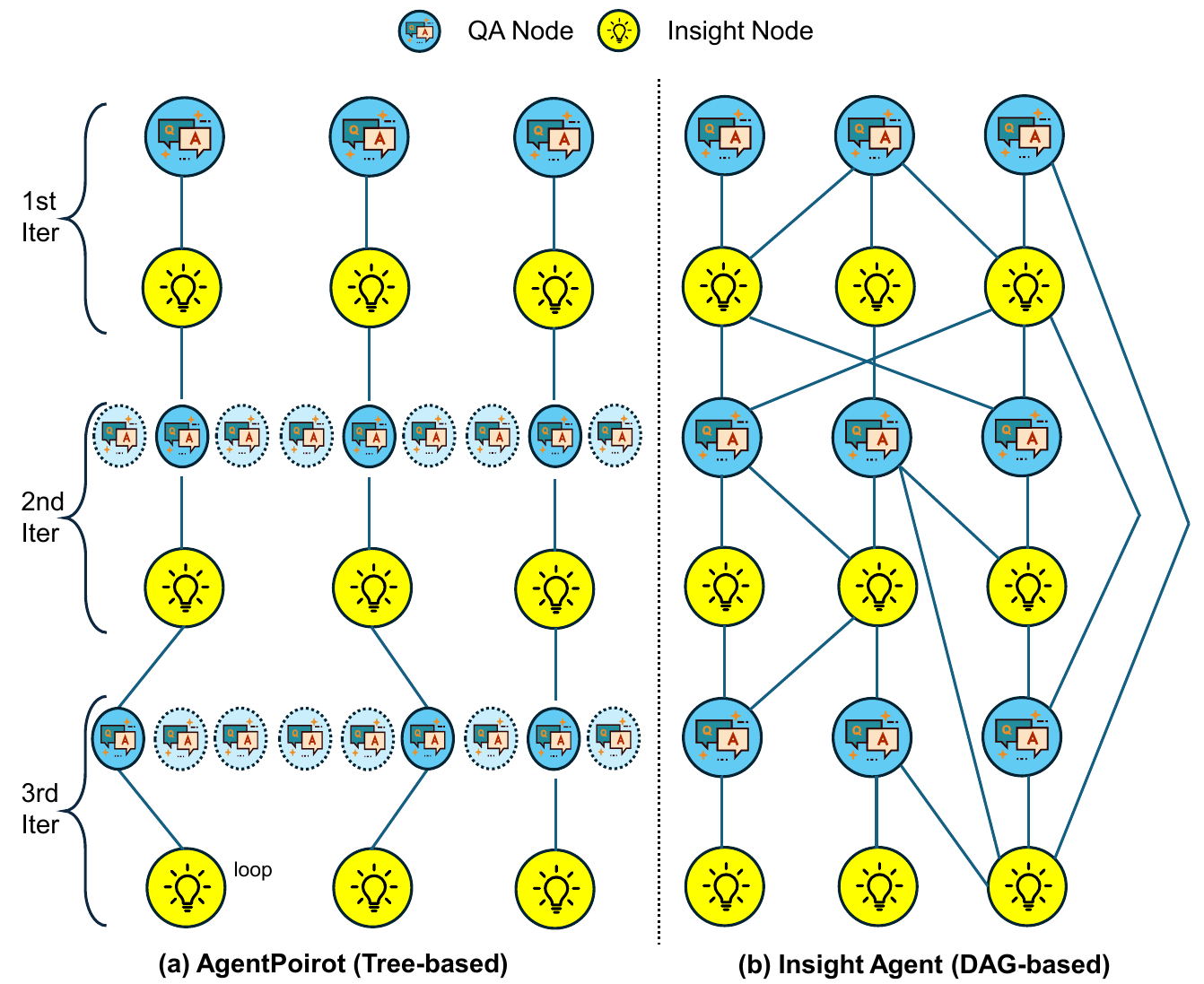}
\caption{Insight generaion process by (a) tree-based process used in AgentPoirot, where each question is selected by LLM from the question candidates and (b) directed acyclic graph-based process used in Insight Agent}
\label{fig:insight_process}
\end{figure}

The Insight Agent operates through an iterative cycle: it generates questions, answers them using Answer Agent, and then synthesizes insights from the resulting QA pairs. The insights generated in one step are then used to inform the question generation in the next, creating a continuous exploratory process.
This process is governed by a Directed Acyclic Graph (DAG) structure, as illustrated in Figure~\ref{fig:insight_process} (b). The graph consists of alternating layers of Question-Answer (QA) nodes and Insight nodes. A new Insight node is generated by synthesizing information from one or more preceding QA nodes, and conversely, a new QA node is generated by drawing upon one or more preceding Insight nodes.
Crucially, a new node can be connected to parent nodes from any previous iteration, not just the immediately preceding one. This DAG structure facilitates the aggregation of multiple lines of inquiry, enabling the generation of more diversified and comprehensive insights compared to a simpler tree-based exploration as in Figure~\ref{fig:insight_process} (a), where insights from different depths and branches are not connected. 
The decision of which nodes to aggregate is determined by the reasoning capabilities of LLM; to guide this process, our prompt explicitly instructs the model to consider synthesizing information from multiple parent nodes when possible.

\subsection{Meta-evaluation of Table Insight}
\label{meta_evaluation}
We conducted a meta evaluation to assess the validity of the evaluation metrics for Table Insight. We sampled 50 pairs of generated insights (Insight A and Insight B) for each ground-truth insight and asked three independent annotators to assess which of the two was closer to the ground truth. Annotators selected one of five relative options: \textbf{A+} (\textit{A is definitely better}), \textbf{A} (\textit{A is slightly better}), \textbf{N} (\textit{A and B are comparable}), \textbf{B} (\textit{B is slightly better}), or \textbf{B+} (\textit{B is definitely better}). Each option was mapped to a normalized score in [-2, -1, 0, 1, 2]. The final human-judgment score for each pair was obtained by averaging the three annotators' scores. In parallel, we computed a metric-derived score based on the difference between the insight-level scores of B and A. We then quantified the agreement between human judgments and the metric-derived scores using both Pearson and Spearman correlations, which yielded coefficients of 0.669 and 0.663, respectively. These results indicate a solid alignment between the proposed metric and human judgments, providing empirical support for the reliability of our evaluation methodology for the Table Insight task.

\begin{table}[t]
\centering
\begin{tabular}{p{0.315\linewidth}|p{0.315\linewidth}|p{0.315\linewidth}}
\toprule
\textbf{GT} & \textbf{Insight Agent} & \textbf{AgentPoirot} \\
\midrule
{\tiny Provinces/Territories with Highest Prevalence: The provinces and territories that exhibited the highest percentages of participant homes testing above the radon guideline were \textbf{Manitoba, New Brunswick}, Saskatchewan, and the \textbf{Yukon}. For example, \textbf{New Brunswick had a raw percentage of 24.8\%} and a population-weighted percentage of 20.6\% of homes above 200 Bq/m³, while Manitoba had 23.7\% raw and 19.4\% population-weighted.} & {\tiny Radon concentrations vary significantly across Canadian provinces and territories, with \textbf{New Brunswick, Yukon, and Manitoba} consistently showing the highest average concentrations and a wider spread of values. \textbf{Notably, New Brunswick has the highest proportion of homes exceeding the 200 Bq/m³ mitigation guideline, with over 25\% of its homes above this threshold}. Within these high-concentration regions, specific health regions and Forward Sortation Areas (FSAs) exhibit even higher localized averages, highlighting the importance of granular regional analysis for targeted mitigation efforts. (score: 0.78)} & {\tiny The average radon concentration in \textbf{New Brunswick} (179.9 Bq/m3) is more than double that of British Columbia (70.98 Bq/m3) and the Northwest Territories (70.96 Bq/m3), highlighting significant regional disparities in radon levels across Canada. (score: 0.37)} \\
\midrule
{\tiny Localized Risk in Provinces with Lower Averages: \textbf{Even in provinces where the overall population-weighted results indicated a lower incidence of homes with elevated radon levels, there were still specific Health Regions with high radon levels}. For example, in \textbf{Ontario}, where the population-weighted estimate was 4.6\% of homes exceeding the guideline, 13 of 36 Health Regions (over one-third) had more than 10\% of homes test above the guideline.} & {\tiny While New Brunswick and Yukon have the highest average radon concentrations, Ontario and Manitoba also show significant radon concerns, particularly regarding high outliers. \textbf{Ontario} has a substantial number of measurements exceeding 500 Bq/m³ (49 instances), and Manitoba has a high proportion of measurements above 200 Bq/m³ (23.67\%), second only to New Brunswick. \textbf{This indicates that even provinces with lower overall average radon concentrations can have localized areas with very high radon levels, necessitating targeted mitigation efforts.} (score: 0.64)} & {\tiny The significant variability in radon concentrations between Health Regions and their provincial averages, as highlighted by the large standard deviation of 40.21 Bq/m3, suggests that localized geological factors or housing characteristics within specific Health Regions may play a more dominant role in radon levels than broader provincial trends. (score: 0.49)} \\
\midrule
{\tiny Age-Related Increases in Mirex and Marker PCBs: Mirex concentrations increased with age in Cycle 1 (2007–2009), with the highest mean serum concentration found in the 60–79 years age group (0.019 ng/g serum) compared to younger groups (e.g., 0.0014 ng/g serum for 6–11 years). \textbf{Similarly, the sum of Marker PCBs (PCB 138, 153, 180) generally showed an increase in concentrations with increasing age across all cycles}, with the 60–79 years age group consistently exhibiting the highest arithmetic means (e.g., 140 ng/g lipid in Cycle 1).} & {\tiny \textbf{Polychlorinated biphenyls (PCBs), particularly 'Marker polychlorinated biphenyls (sum of PCB 138, 153, 180)', show a clear age-related accumulation}, with significantly higher average measured amounts in older age groups (40-79 years) compared to younger ones (3-39 years). This suggests a persistent presence and bioaccumulation of these substances over a person's lifetime. (score: 0.71)} & {\tiny For Polychlorinated biphenyls, the AM-MA values for the 'Total' gender group consistently increase with age, with the 60-79 age group showing AM-MA values as high as 9.62, significantly higher than the 12-19 age group which has values as low as 0.89. (score: 0.31)} \\
\bottomrule
\end{tabular}
\caption{Apple-to-apple comparison among GT insight, insight from Table Insight, and one from AgentPoirot.}
\label{tab:a2a_comparison}
\end{table}

\end{document}